\documentclass[conference]{IEEEtran}
\IEEEoverridecommandlockouts

\usepackage{cite}
\usepackage{amsmath,amssymb,amsfonts}
\usepackage{algorithmic}
\usepackage{graphicx}
\usepackage{booktabs} 
\usepackage{textcomp}
\usepackage{xcolor}
\def\BibTeX{{\rm B\kern-.05em{\sc i\kern-.025em b}\kern-.08em
    T\kern-.1667em\lower.7ex\hbox{E}\kern-.125emX}}

\makeatletter
\newcommand{\linebreakand}{%
  \end{@IEEEauthorhalign}
  \hfill\mbox{}\par
  \mbox{}\hfill\begin{@IEEEauthorhalign}
}
\makeatother

\begin{document}

\title{RAG-BioQA: A Retrieval-Augmented Generation Framework for Long-Form Biomedical Question Answering}

\author{
    \IEEEauthorblockN{Lovely Yeswanth Panchumarthi}
    \IEEEauthorblockA{\textit{Department of CSE} \\
    \textit{Emory University}\\
    Georgia, USA \\
    lpanch2@emory.edu}
    \and
    \IEEEauthorblockN{Sumalatha Saleti}
    \IEEEauthorblockA{\textit{Department of CSE} \\
    \textit{SRM University AP}\\
    Andhra Pradesh, India \\
    sumalatha.s@srmap.edu.in}
    \and
    \IEEEauthorblockN{Sai Prasad Gudari}
    \IEEEauthorblockA{\textit{Department of CSE} \\
    \textit{Emory University}\\
    Georgia, USA \\
    sgudari@emory.edu}
    \linebreakand 
    \IEEEauthorblockN{Atharva Negi}
    \IEEEauthorblockA{\textit{Department of CSE} \\
    \textit{Emory University}\\
    Georgia, USA \\
    anegi2@emory.edu}
    \and
    \IEEEauthorblockN{Praveen Raj Budime}
    \IEEEauthorblockA{\textit{Department of CSE} \\
    \textit{Trine University}\\
    Michigan, USA \\
    Budimepraveenraj@gmail.com}
    \and
    \IEEEauthorblockN{Harsit Upadhya}
    \IEEEauthorblockA{\textit{Department of CSE} \\
    \textit{Emory University}\\
    Georgia, USA \\
    hupadhy@emory.edu}
}

\maketitle

\begin{abstract}
The rapidly growth of biomedical literature creates challenges acquiring specific medical information. Current biomedical question-answering systems primarily focus on short-form answers, failing to provide comprehensive explanations necessary for clinical decision-making. We present RAG-BioQA, a retrieval-augmented generation framework for long-form biomedical question answering. Our system integrates BioBERT embeddings with FAISS indexing for retrieval and a LoRA fine-tuned FLAN-T5 model for answer generation. We train on 181k QA pairs from PubMedQA, MedDialog, and MedQuAD, and evaluate on a held-out PubMedQA test set. We compare four retrieval strategies: dense retrieval (FAISS), BM25, ColBERT, and MonoT5. Our results show that domain-adapted dense retrieval outperforms zero-shot neural re-rankers, with the best configuration achieving 0.24 BLEU-1 and 0.29 ROUGE-1. Fine-tuning improves BERTScore by 81\% over the base model. We release our framework to support reproducible biomedical QA research.
\end{abstract}


\section{Introduction}
The biomedical domain is currently witnessing an exponential growth in literature, with databases such as PubMed indexing millions of citations and adding thousands of new articles daily \cite{yoo2012data,masic2012line}. This profusion of data presents a "paradox of abundance" for healthcare professionals, researchers, and clinical decision-makers: while the volume of available knowledge is vast, the ability to efficiently locate, synthesize, and apply specific information to clinical scenarios is increasingly constrained by human cognitive limits \cite{sandhu2006clinical}. Lists of documents are returned by conventional search engines. The information must be manually read and synthesized by users. This procedure takes a long time. In clinical settings when quick choices are required, it is not practicable \cite{natarajan2010analysis, sivarajkumar2024clinical, fok2024qlarify}. Confusion is caused by large datasets. We urgently require precise responses. A solution is offered by automated Q\&A systems. These tools provide precise information by filtering out noise. Knowledge is accessible without the need for human search \cite{hoffner2017survey, rogers2023qa}.   

The limitations of general tools are revealed by biomedical tasks. We come across challenging terminology. Conventional definitions frequently fall short. We have to monitor particular causes and consequences. Facts are essential to patient safety. Mistakes put lives in danger. We require complete accuracy \cite{jin2022biomedical, li2024developing}. There are limitations to the tools available today. They extract particular passages from the text. We get brief responses. We receive a straightforward "yes" or "no." We locate single names \cite{li2024pedants, wang2023survey}. These systems are helpful for factoid queries, but they are unable to meet the complex information needs of clinicians who need lengthy, explanatory responses that include reasoning, context, and supporting data \cite{currie2003clinical}. 

The primary drawback of existing Large Language Models (LLMs) in healthcare—the propensity to "hallucinate," or produce plausible but factually incorrect information—is the driving force behind this study \cite{huang2025survey, cossio2025comprehensive, kim2025medical}. In a clinical context, a hallucinated drug interaction or an invented treatment guideline can have catastrophic consequences for patient safety \cite{johnson2008human, breggin2007brain, sivanesan2016opioid}. Retrieval-Augmented Generation (RAG) offers a promising solution by grounding generative models in external, non-parametric memory (e.g., verified medical literature) \cite{neha2025retrieval, oche2025systematic}. By retrieving relevant documents and conditioning the generation on this evidence, RAG systems aim to combine the fluency of LLMs with the reliability of retrieval systems.   

RAG systems face gaps in biomedical use. Developers optimize existing frameworks for open domain tasks. Neural re-ranking models like ColBERT \cite{khattab2020colbert} and MonoT5 \cite{nogueira2020document} follow this trend. These models depend on datasets such as MS MARCO  \cite{nguyen2016ms}. There are fewer tools designed for medical data. In medical settings, re-ranking techniques are difficult. Meaning cannot be captured by general lexical overlap. To assess relevance, we require exact medical ontology. This transferability has not been sufficiently investigated by researchers \cite{frisoni2021survey}. Reranking techniques utilising neural networks require significant computational power. In many clinical settings, resources are scarce. The high expense involved in purchasing these tools prevents their use in hospitals. We investigate a novel question answering framework applicable to the retrieval of long answers within the biomedical domain. We are currently looking into the complexities and costs associated with retrieval. The quality of the generated text is evaluated. The research reveals the kind of compromises which would have to be made for the test to be used by clinicians:

This investigation is driven by three primary research questions (RQs) designed to dissect the efficacy of RAG components in the biomedical context:

\begin{itemize}
    \item RQ1: Can Retrieval-Augmented Generation (RAG) effectively improve the quality of long-form biomedical question answering compared to non-retrieval baselines?
    \item RQ2: We investigate the effectiveness of three different re-ranking strategies: sparse (BM25 \cite{robertson2009probabilistic}), dense interaction (ColBERT \cite{khattab2020colbert}) and sequence to sequence (MonoT5 \cite{nogueira2020document}) for medical queries.
    \item RQ3: What is the impact of LoRA on the performance of the generative model in domain adaptation when fine-tuning with Parameter-Efficient Fine-Tuning (PEFT) \cite{ding2023parameter}?
\end{itemize}

The use of a search engine like PubMed \cite{chen2004content} to supplement a generative model in answering medical questions is investigated. The goal is to determine whether this method results in answers that are both more coherent and factually accurate than those provided by a traditional AI system. In this work, we determine whether a zero-shot re-ranker can outperform a standard dense retrieval system which has been adapted to the domain through the use of BioBERT \cite{lee2020biobert}. This text highlights a problem which arises from the disparity between the type of data which a neural information retrieval system has been trained upon and the type of data which it is being asked to search through. An experiment is reported on adapting a pre-trained language model, Flan-T5 \cite{khattab2020colbert}, to the field of medicine by applying the Low-Rank Adaptation (LoRA) technique. The model's efficiency and text generation abilities are assessed.

\section{Related Work}
Biomedical question answering has changed. Experts moved from rule-based systems to deep neural networks. Now we use retrieval-augmented paradigms. This section reviews four areas. We examine Biomedical QA and RAG methodologies. We evaluate Long-form Generation and Re-ranking strategies.

The development of mathematical models in biomedical science reveals a conflict of philosophies with regard to their use. Training from scratch on 14 million abstracts yields the best results, according to PubMedBERT \cite{gao2021rethink}. This method outperformed the mixed domain training methodology. The SciBERT \cite{beltagy2019scibert} model has been pre-trained on a massive corpus of 1.14 million scientific papers. BioLinkBERT \cite{yasunaga2022linkbert} showed that using citation links could improve MedQA-USMLE \cite{jin2021disease} scores by seven percent. These developments have given a boost to the field. Biomedical language model BioGPT \cite{luo2022biogpt} used 1.5 billion parameters to reach a score of 81 percent on PubMedQA. The BioMedLM \cite{bolton2024biomedlm} model has been scaled up to 2.7 billion parameters. We obtain a custom word tokeniser for medical terms. The model performed at 57.3\% on MedQA in its compact form.

The foundational RAG architecture established the paradigm of combining parametric memory with non-parametric retrieval, explicitly noting that systems could be "endowed with a medical index" for domain-specific applications. Architectural innovations have substantially improved retrieval generation integration. Fusion-in-Decoder \cite{yu2022kg} addressed the computational bottleneck of processing multiple passages by encoding them independently before fusion in the decoder achieving 51.4\% exact match on Natural Questions while scaling to 100 retrieved passages with linear rather than quadratic complexity. RETRO \cite{bhuiyan2025retrieval} introduced chunked cross-attention that retrieves for each 64-token segment from a 2 trillion token database, matching GPT-3 (175B) \cite{floridi2020gpt} performance with 25× fewer parameters. Atlas demonstrated that joint pre-training of retriever and generator enables remarkable few shot capability outperforming 540B PaLM \cite{chowdhery2023palm} by 3\% on Natural Questions using only 64 examples.

Adaptive retrieval mechanisms represent the current frontier. Self-RAG trained models to decide when to retrieve using special reflection tokens enabling on-demand retrieval and self critique of generated content. FLARE \cite{jiang2023active} implemented forward-looking active retrieval that triggers on low-confidence tokens, particularly effective for multi-hop reasoning in long-form generation. REPLUG \cite{shi2024replug} demonstrated that even black box LLMs can benefit from retrieval through parallel document prepending and probability ensembling applicable to closed source medical AI deployments. MedRAG \cite{xiong2024benchmarking} provide systematic evaluation across 7,663 questions from five medical QA datasets, demonstrating that combining multiple corpora (PubMed, StatPearls, textbooks) with diverse retrievers achieves up to 18\% accuracy improvement and elevates GPT-3.5 to GPT-4-level performance. 

Long-form question answering presents distinct challenges for biomedical applications where comprehensive explanations must synthesize evidence from multiple sources. The ELI5 dataset \cite{fan2019eli5} established benchmarks for explanatory QA with 270,000 questions requiring ~130-word answers synthesized from 100 web documents. However, attribution mechanisms have become essential for medical QA credibility. WebGPT \cite{nakano2021webgpt} pioneered interactive web browsing with snippet level citations, trained via reinforcement learning from human feedback on 6,000 demonstrations. GopherCite advanced this with verifiable quoted snippets from Google Search results, achieving 80\% high-quality responses. 

Generalization has emerged as the critical differentiator for medical applications. TAS-B achieved best zero shot generalization among dense retrievers through balanced topic-aware sampling and dual teacher distillation trainable on a single consumer GPU while outperforming BM25 by 44\% and prior models by 5\%. Contriever enabled unsupervised dense retrieval through contrastive learning with random cropping augmentation, competitive with BM25 on 11/15 BEIR datasets without any labeled training data. E5 embeddings became the first model to consistently beat BM25 under zero shot BEIR evaluation through weakly-supervised contrastive pre-training on 270 million text pairs.

\section{Methodology}

\subsection{Framework Overview}
\begin{figure}
    \centering
    \includegraphics[width=1\linewidth]{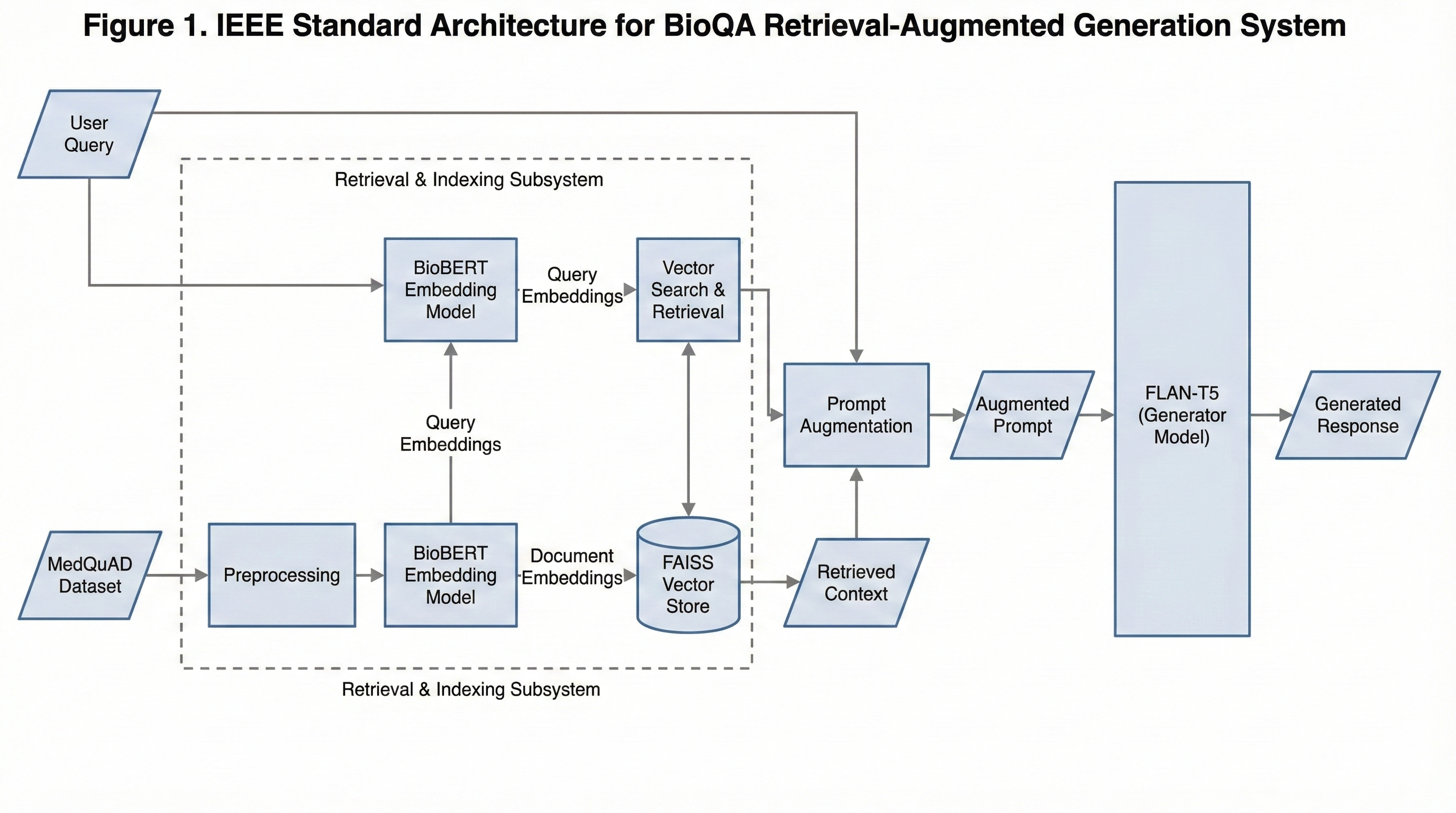}
    \caption{The RAG-BioQA framework follows a clear process. The system retrieves contexts using BioBERT embeddings and FAISS indexing. We apply re-ranking strategies to select informative data. The system combines these contexts with our query. A fine-tuned T5 model generates comprehensive answers.}
    \label{fig:reranker_framework}
\end{figure}

RAG-BioQA operates in two stages: (1) retrieval of relevant QA pairs from a curated biomedical database, and (2) answer generation conditioned on retrieved examples. The main distinction between RAGs utilised for document retrieval and this question answering RAG is that it retrieves question and answer pairs in their entirety, thereby providing the model with the contexts and the structure of the given answers. Figure \ref{fig:reranker_framework} illustrates the architecture.

\subsection{Preprocessing Pipeline}
We make use of pre-processed question and answer pairs from the PubMedQA dataset within our retrieval system. Most articles do not cite raw biomedical literature. This layout is structured to fit in with what we already know. This tool synthesizes evidence more quickly than full document retrieval systems. The pipeline includes these steps:

\begin{enumerate}
    \item \textbf{Data Filtering and Cleaning: }For training and evaluation, we use the PubMedQA dataset after it has been filtered. We eliminate low-quality annotations. Standardising medical abbreviations puts their meanings into ordinary English. Biomedical terminology is processed in this system for accurate meaning.
    
    \item \textbf{Embedding Generation:} For each query in our dataset, we use BioBERT to generate a dense vector representation of that query along with its related passage. It employs a training regime on biomedical literature initially. We apply mean pooling over the token embeddings. This results in fixed-length vectors which are suitable for similarity searching.
    
    \item \textbf{Index Construction:} We use the FAISS \cite{singh2023analyzing} library to index the output vector space. This application performs high-performance similarity searches on vectorial datasets. In the implementation, an L2 index is used for retrieval of the nearest neighbour with an exact match. Using our existing database allows the retrieval of the most accurate results.
\end{enumerate}

This processing stage results in a data base structured of question context pairs with the vectors associated with each, enabling fast retrieval during the question answering process \cite{trangcasanchai2024improving}.

\subsection{Retrieval Module with Re-ranking Strategies}
Embedding Generation. We encode all QA pairs using BioBERT-base-cased-v1.1, a BERT model pre-trained on PubMed abstracts and PMC full-text articles \cite{gupta2023top}. For each QA pair, we concatenate the question and answer, pass through BioBERT, and apply mean pooling over token embeddings to obtain a 768-dimensional representation.

Index Construction. We index all QA pair embeddings using FAISS with an IndexFlatL2 configuration for exact nearest neighbor search. At query time, the input question is embedded using the same BioBERT model, and we retrieve the top-k most similar QA pairs (k=16).

Re-ranking Strategies. We compare four strategies for selecting the final n=4 contexts:

\begin{itemize}
    \item \textbf{FAISS (baseline):} Use L2 distance scores directly without re-ranking.
    \item \textbf{BM25:} Re-rank by lexical overlap using term frequency and inverse document frequency.
    \item \textbf{ColBERT:} Re-rank using late interaction between query and document token embeddings.
    \item \textbf{MonoT5:} Re-rank using a sequence-to-sequence model trained for relevance classification.
\end{itemize}

BM25, ColBERT, and MonoT5 are applied zero-shot without biomedical fine-tuning.

\subsection{Generation Module}
The answer generation module synthesizes retrieved contexts. We produce long-form answers to biomedical questions. We fine-tune T5 models for this task. These models show strong performance in natural language generation.

\begin{enumerate}
    \item \textbf{Context Integration:} We concatenate the retrieved contexts and the query question into a unified input format. This structure provides the model with all necessary evidence. We use specific markers to separate each context from the question.
    
    \texttt{Context: [Retrieved QA Pair 1] [Retrieved QA Pair 2] ... [Retrieved QA Pair n] Question: [Query] Answer:}
    
    We format each retrieved pair as \texttt{Question: [Retrieved Question] Answer: [Retrieved Answer]}. This structure provides the model with the question and the corresponding answer. We use this format to establish clear context.
    
    \item \textbf{Model Fine-tuning:} We fine-tune a pre-trained FLAN-T5-base model on the formatted training data. We implement Parameter-Efficient Fine-Tuning using Low-Rank Adaptation to improve efficiency. This method reduces memory requirements. We update a small subset of model parameters. Performance remains stable.
    
    \item \textbf{Answer Generation:} At inference time, the pre-processed input to the models is passed through the fine-tuned T5 model. The model generates a long-form answer. The information gathered from the retrieved sources is integrated here. We perform decoding using a beam search technique of size 4. In order to encourage response detail, we apply length normalization.
\end{enumerate}

\section{Experimental Setup}

\subsection{Dataset}
The training data set is constructed by combining the MedQuAD, MedDialog and PubMed question-answering data sets. The dataset contains 181,488 Q A pairs. The content covers clinical questions and patient-doctor dialogues. Our medical reference section covers a wide range of subjects. The data was divided into three groups. We use 70\% for training. We assign 15\% to validation. We reserve 15\% for testing. We keep the PubMedQA test set for the final evaluation. We allow no overlap between training and test pairs. PubMedQA provides expert-annotated questions. These questions link to article abstracts. This structure supports long-form biomedical answer evaluation.

\subsection{Implementation Details}
Our implementation uses the following technologies and parameters:

\begin{itemize}
    \item \textbf{Embedding Model:} BioBERT-base-cased-v1.1 (768-dimensional embeddings).
    \item \textbf{Retrieval:} FAISS IndexFlatL2 with $k=16$ initial candidates, n=4 final contexts.
    \item \textbf{Re-ranking:} BM25 (rank-bm25 library), ColBERT (official checkpoint), MonoT5 (castorini/monot5-base-msmarco).
    \item \textbf{Generator:} FLAN-T5-base (250M parameters) fine-tuned with LoRA
    \item \textbf{LoRA Configuration:} rank $r=16$, alpha $\alpha=32$, dropout $p=0.1$
    \item \textbf{Training:} Batch size of 8, learning rate of 5e-5, 3 epochs
    \item \textbf{Hardware:} NVIDIA A100 GPUs with 128GB memory
\end{itemize}

\subsection{Baselines}
We compare our approach against the following baselines:

\begin{itemize}
    \item \textbf{Base T5 + FAISS:} Pre-trained FLAN-T5 without fine-tuning, using FAISS retrieval.
    \item \textbf{Finetuned T5 + FAISS:} LoRA fine-tuned FLAN-T5 with FAISS retrieval (no re-ranking).
    \item \textbf{Finetuned T5 + [Re-ranker]:} Fine-tuned model with BM25, ColBERT, or MonoT5 re-ranking
\end{itemize}

This setup isolates the impact of fine-tuning by comparing Base models against Finetuned models. We also isolate re-ranking strategies by comparing FAISS against other methods. These comparisons show the performance gain from domain adaptation. We measure how each component contributes to answer quality.

\section{Evaluation Metrics}
We evaluate generated answers using four complementary metrics:

\begin{enumerate}

\item \textbf{BLEU-1 (Bilingual Evaluation Understudy - Unigram Precision): } This sentence could be measured using the BLEU-1 score which is a metric to check the precision of a system's output with respect to single words.

\begin{equation}
    \text{BLEU-1} = \text{BP} \cdot \exp\left( \log p_1 \right)
\end{equation}

\begin{equation}
    p_1 = \frac{\sum_{\text{word} \in \text{generated}} \min(\text{count}_{\text{gen}}(\text{word}), \text{count}_{\text{ref}}(\text{word}))}{\sum_{\text{word} \in \text{generated}} \text{count}_{\text{gen}}(\text{word})}
\end{equation}

\begin{equation}
    \text{BP} = \begin{cases} 
      1 & \text{if } c > r \\
      e^{(1 - \frac{r}{c})} & \text{if } c \leq r
   \end{cases}
\end{equation}

\begin{itemize}
  \item \( c \): length of the generated sentence
  \item \( r \): length of the reference sentence
\end{itemize}

\item \textbf{ROUGE-1 (Recall-Oriented Understudy for Gisting Evaluation - Unigram Recall): } ROUGE-1 measures the co-occurrence of one word in a system output versus a reference summary. The number of words in the output is 16 while in the reference is 18. The measure of precision refers to the number of correct predictions out of all the predictions made by the system.

\begin{equation}
    \text{ROUGE-1} = \frac{\sum_{\text{word} \in \text{ref}} \min(\text{count}_{\text{ref}}(\text{word}), \text{count}_{\text{gen}}(\text{word}))}{\sum_{\text{word} \in \text{ref}} \text{count}_{\text{ref}}(\text{word})}
\end{equation}

\item \textbf{BERTScore (Precision): } In the BERTScore the similarity between tokens is evaluated by utilising contextual embeddings. The precision metric evaluates the extent to which the reference sentences match sentences generated by the system. The evaluation of the meaning of the query and the content is based on the conceptual similarity of the two.

\begin{equation}
    \text{BERTScore}_{\text{Precision}} = \frac{1}{|X|} \sum_{x_i \in X} \max_{y_j \in Y} \text{cosine}(x_i, y_j)
\end{equation}

\begin{equation}
    \text{cosine}(x_i, y_j) = \frac{x_i \cdot y_j}{\|x_i\| \|y_j\|}
\end{equation}

\begin{itemize}
  \item \( X \): token embeddings of the generated sentence
  \item \( Y \): token embeddings of the reference sentence
\end{itemize}

\item \textbf{METEOR (Metric for Evaluation of Translation with Explicit ORdering): } This aligns the reference and model sentences by positions to evaluate the model's performance. The search is performed on word stems as well as the words themselves. We include synonym and paraphrase matches. This measure allows for a relative assessment of linguistic equivalence.

\begin{equation}
    \text{METEOR} = F_{\text{mean}} \cdot (1 - \text{Penalty})
\end{equation}

\begin{equation}
    F_{\text{mean}} = \frac{10 \cdot P \cdot R}{R + 9P}
\end{equation}

\begin{equation}
    P = \frac{m}{w_{\text{gen}}}, \quad R = \frac{m}{w_{\text{ref}}}
\end{equation}

\begin{equation}
    \text{Penalty} = \gamma \cdot \left(\frac{ch}{m}\right)^{\theta}
\end{equation}

\begin{itemize}
  \item \( m \): number of mapped unigrams (matches)
  \item \( w_{\text{gen}} \): total unigrams in the generated sentence
  \item \( w_{\text{ref}} \): total unigrams in the reference sentence
  \item \( ch \): number of chunks (contiguous matched subsequences)
  \item \( \gamma \), \( \theta \): tunable parameters (commonly \( \gamma = 0.5, \theta = 3 \))
\end{itemize}
\end{enumerate}

\section{Results and Discussion}

\begin{table}[ht]
\centering
\caption{Performance comparison of different model configurations on the PubMedQA test set.}
\label{tab:results}
\resizebox{\columnwidth}{!}{%
\begin{tabular}{lcccc}
\toprule
\textbf{Model} & \textbf{BLEU-1} & \textbf{ROUGE-1} & \textbf{BERTScore} & \textbf{METEOR} \\
\midrule
Base T5 + FAISS & 0.2065 & 0.2618 & 0.1132 & 0.1948 \\
Finetuned T5 + FAISS & \textbf{0.2415} & \textbf{0.2918} & \textbf{0.2054} & \textbf{0.2264} \\
Finetuned T5 + BM25 & 0.2221 & 0.2714 & 0.1318 & 0.2054 \\
Finetuned T5 + ColBERT & 0.2218 & 0.2713 & 0.1364 & 0.2053 \\
Finetuned T5 + MonoT5 & 0.2172 & 0.2632 & 0.1277 & 0.2023 \\
\bottomrule
\end{tabular}%
}
\end{table}

\subsection{Impact of Fine-tuning}
Fine-tuning yields consistent improvements across all metrics refer \ref{tab:results}. Comparing Base T5 + FAISS to Finetuned T5 + FAISS, we observe a 17\% improvement in BLEU-1 (0.2065 → 0.2415), 11\% in ROUGE-1 (0.2618 → 0.2918), and 81\% in BERTScore (0.1132 → 0.2054). The substantial BERTScore gain indicates that fine-tuning improves semantic alignment with reference answers, not merely lexical overlap. This confirms that domain adaptation through LoRA effectively specializes the generator for biomedical content.

\subsection{Impact of Re-ranking Strategies}
A notable finding is that FAISS retrieval without re-ranking outperforms all re-ranking strategies. The fine-tuned model with FAISS achieves the highest scores across all metrics, while BM25, ColBERT, and MonoT5 re-ranking each degrade performance. We attribute this to domain mismatch. ColBERT and MonoT5 are trained on MS MARCO, a general-domain retrieval dataset. Biomedical text contains specialized terminology, abbreviations, and semantic relationships absent from general web queries. When applied zero-shot, these re-rankers may prioritize lexical patterns learned from general text that do not transfer to biomedical contexts.

BM25 performs slightly better than the neural re-rankers but still underperforms FAISS. This suggests that lexical matching alone is insufficient for biomedical retrieval, where semantically related terms (e.g., "myocardial infarction" and "heart attack") may share few surface tokens. In contrast, BioBERT embeddings are pre-trained on PubMed and PMC corpora, encoding biomedical semantics directly. This domain alignment appears more valuable than the sophisticated matching mechanisms of neural re-rankers applied without domain adaptation.

\subsection{Analysis of Re-ranker Performance}
Among the re-ranking strategies, performance differences are small. BM25 achieves marginally higher scores than ColBERT and MonoT5, though all three perform comparably. This suggests that in the absence of domain adaptation, the choice of re-ranking algorithm matters less than the domain gap itself.
The consistent underperformance of neural re-rankers challenges findings from general IR research, where re-ranking typically improves retrieval quality. Our results suggest this assumption does not hold in specialized domains without appropriate fine-tuning.

\subsection{Efficiency Considerations}
Beyond accuracy, FAISS retrieval offers practical advantages. Dense retrieval with FAISS requires only a single forward pass through BioBERT and a nearest neighbor search. Re-ranking with ColBERT or MonoT5 requires additional model inference for each candidate, increasing latency and compute cost. Given that re-ranking provides no accuracy benefit in our setting, FAISS offers better accuracy-efficiency tradeoff for biomedical QA.

\section{Conclusion}
We presented RAG-BioQA, a retrieval-augmented generation framework for long-form biomedical question answering. Our system combines BioBERT embeddings for dense retrieval, FAISS indexing for efficient search, and a LoRA fine-tuned FLAN-T5 model for answer generation. We train on 181k QA pairs from three biomedical datasets and evaluate on PubMedQA.
Our experiments yield two main findings. First, domain-specific fine-tuning substantially improves generation quality, with BERTScore improving by 81\% over the base model. Second, domain-adapted dense retrieval (BioBERT + FAISS) outperforms zero-shot neural re-rankers (BM25, ColBERT, MonoT5). This suggests that for biomedical QA, investing in domain-aligned embeddings is more effective than applying sophisticated re-ranking methods without domain adaptation.

\section{Limitations}
In comparison with general information retrieval, our experiments yield somewhat different results from those which have been seen previously, as re-ranking does not improve retrieval effectiveness. This gap in research shows how significant the difference between the fields of the lab and the clinical setting is. In comparison to other models such as Med-PaLM and GPT-4, our approach has the disadvantage of being less capable and of a smaller size. Our approach offers the advantages of open source software, maintains user data privacy by local processing and also provides results which can be easily verified. BioBERT-FAISS sets a high benchmark for publicly available biomedical Retrieval-Augmented Generation models in terms of ROUGE-1 score which is 0.29.

\begin{itemize}
    \item \textbf{Hallucinations:} The model still exhibited the problem of hallucinated certainty as found in the qualitative assessment. This happens when tentative results of an exploratory study are described in a definitive way. Such a mislabelling can be found particularly in clinical contexts.
    \item \textbf{Context Window:} Due to the 512 token context window of T5, we limited our retrieval to the top 4 documents. Forcing the researcher to browse through the entire document to identify relevant evidence may force the exclusion of relevant evidence if the documents are long.
    \item \textbf{Metric Limitations: }Automated evaluation metrics such as BLEU ratio do not assess the medical correctness of responses produced by a machine. A low quality answer, medically incorrect, may have a high score in this system if it contains a lot of the same words as the correct summary.
\end{itemize}

\bibliographystyle{IEEEtran}
\bibliography{references}

\end{document}